\documentclass{article} 

\newcommand{\mr}[1]{\mathrm{#1}}
\newcommand{\mb}[1]{\mathbf{#1}}
\newcommand{\mt}[1]{\mathtt{#1}}
\newcommand{\mc}[1]{\mathcal{#1}}
\newcommand{\+}{~\&~}
\newcommand{\?}{~?~}
\newcommand{\bv}{\mathbf{v}}
\newcommand{\SL}{\textsc{S-Lang}}
\newcommand{\SN}{\textsc{S-Net}}
\newcommand{\SR}{\textsc{S-Rep}}
\newcommand{\ignore}[1]{}
\newcommand{\subsectionPH}[1]{\emph{#1}.}

\usepackage{iclr2018_workshop,times}
\usepackage{hyperref}
\usepackage{url}
\usepackage{graphicx}    

\title{Learning and analyzing vector encoding of symbolic representations}

\author{Roland Fernandez, Asl{\i} \c{C}eliky{\i}lmaz, Rishabh Singh\thanks{Currently at Google Brain, Mountain View, CA, USA.} \\
Microsoft Research AI\\
Redmond, WA 98052, USA \\
\texttt{\{rfernand,aslicel\}@microsoft.com}, \texttt{rishabh.iit@gmail.com}  \\
\And
Paul Smolensky\\
Microsoft Research AI \& Johns Hopkins University \\
Redmond, WA 98052, USA \& Baltimore, MD 21218, USA \\
\texttt{psmo@microsoft.com, smolensky@jhu.edu} \\
}

\begin{document}

\maketitle

\begin{abstract}
We present a formal language with expressions denoting general symbol structures and queries which access information in those structures.
A sequence-to-sequence network processing this language learns to encode symbol structures and query them.
The learned representation (approximately) shares a simple linearity property with theoretical techniques for performing this task. 
\end{abstract}

\vspace{-.2in}

\section{Overview: \SL, \SN\ and \SR}
\label{sec:overview}

Embedding words and sentences in vector spaces has brought many symbolic tasks (especially in NLP) within the scope of deep neural network (DNN) models \citep{Hinton88, Palangi16deep, Pollack90recursive, Socher10learning, Weston15}.
In general, DNNs may be expected to benefit if they can incorporate some of the power of symbolic computation without compromising the power of deep learning.
The problem of embedding general symbol structures in vector spaces, and performing symbolic computation with these vectors, has been addressed theoretically, but these methods can require very large embedding spaces --- e.g., Tensor Product Representation, TPR \citep{Lee16, Smolensky90tensor, Smolensky06harmonic} --- or major error-correction/clean-up processes --- e.g., Holographic Reduced Representation, HRR \citep{Crawford16, Plate93holographic, Plate02, Plate03holographic} (Sec.~\ref{sec:superTheo}; see also \citet{Kanerva09, Touretzky90}).
We show here (Sec.~\ref{sec:results}) that deep learning can itself discover satisfactory methods of embedding general symbol structures, methods that operate in relatively small vector spaces without need for clean-up processing.

We define a general formal language scheme in which expressions denote symbol structures.
Such a language will be called an \SL\ (Sec.~\ref{sec:task}).
Information within these structures is accessed by evaluating query expressions within the language.
The model that learns to encode structure-denoting expressions and to evaluate queries over these structures \citep{Zaremba14} is a simple bidirectional encoder-decoder model that operates on symbols in the formal language one at a time \citep{Cho14}.
We call such a model an \SN\ (Sec.~\ref{sec:model}),
and call the vector embedding of an \SL\ learned by an \SN\  an \SR.

Our \SN\ learns to evaluate \SL\ expressions with a high degree of accuracy (Sec.~\ref{sec:results}).
Furthermore, upon analysis, the \SR\ that \SN\ learns turns out to exhibit a simple linearity property --- the Superposition Principle --- that is crucial to both theoretical models, TPR and HRR (Sec.~\ref{sec:superpos}).

\section{The task: Embedding general symbolic structures in vector spaces and accessing their contents}
\label{sec:task}

\begin{figure}[h]
\begin{center}
\vspace{-7.8cm}
\includegraphics[width=1.0\textwidth]{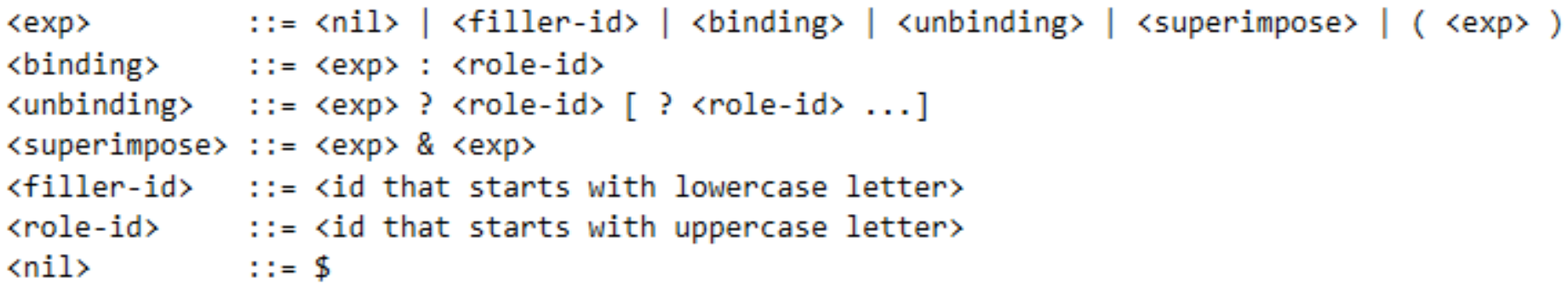}
\vspace{-8.5cm}
\caption{Grammar of \SL: expressions denoting symbol structures and access queries.}
\label{fig:grammar}
\end{center}
\end{figure}

\subsectionPH{Symbol-structure-denoting expressions: Structural-role binding}
\label{sec:expr}
In general, a symbol structure can be characterized as a set of symbols each bound to a role that it plays in the structure \citep[141]{Newell80physical}.
The method is applicable to any type of symbol structure, but we focus on binary trees here.
The simple binary tree $\mc{T} = [\mt{a}~[\mt{b}~\mt{c}]]$ consists of the symbols $\mt{a}, \mt{b}, \mt{c}$ respectively bound to the roles $\mt{R_{0}}, \mt{R_{01}, \mt{R_{11}}}$, where $\mt{R_{01}}$ is the role of left-child (`0') of right-child (`1') of root, etc.
Symbolic structural roles are typically recursive.
The recursive character of binary tree roles can be seen by viewing $\mt{R_{01}}$ as the symbol $\mt{R_{0}}$ bound to the role $\mt{R_{1}}$. 
In the simple formal language we develop here, \SL, the tree $\mc{T}$ will be denoted by the expression $\mt{a\!:\!L \+ b\!:\!L\!:\!R \+ c\!:\!R\!:\!R}$, where $\mt{L}, \mt{R}$ respectively abbreviate the roles $\mt{R_{0}, R_{1}}$. 
The grammar of \SL\ is shown in Fig.~\ref{fig:grammar}. 

\subsectionPH{Query-denoting expressions: Structural-role unbinding}
\label{sec:queries}
A minimal requirement for a vector embedding a symbol structure is that it be possible to extract, with vector computation, (the embedding of) the symbol that is bound to any specified role.
In \SL\ (see Fig.~\ref{fig:grammar}), the query denoted $\mt{?}~\mt{S}$ asks for the structure (possibly a single symbol), bound to the role denoted $\mt{S}$; this is role-unbinding.
Thus the expression $\mt{a\!:\!L \+ b\!:\!L\!:\!R \+ c\!:\!R\!:\!R ~?~ L}$ asks for the structure filling the role $\mt{L}$ in $\mc{T}$, i.e., the structure forming the left child of the root, which in $\mc{T}$ happens to be a single symbol, $\mt{a}$.
Similarly, the expression $\mt{a\!:\!L \+ b\!:\!L\!:\!R \+ c\!:\!R\!:\!R ~?~L\!:\!R}$ asks for the left child of the right child of the root, and so has the value $\mt{b}$.
An \SL\ query can return a structure rather than a single symbol. 
The expression $\mt{a\!:\!L \+ b\!:\!L\!:\!R \+ c\!:\!R\!:\!R ~?~ R}$ asks for the right child of the root of $\mc{T}$, which is the structure $\mt{b\!:\!L \+ c\!:\!R}$, denoting $\mc{T}$'s right sub-tree, $\mt{[ b ~ c ]}$.

\subsectionPH{Expressions combining querying and structure-building}
\label{sec:genExpr}
The general expression in \SL\ allows structure that is returned by queries to be used to build new structures.
Table~\ref{tbl:exs} provides examples of expressions correctly evaluated by \SN.

\begin{table}[t]
\vspace{-.4cm}
\caption{Example input/output pairs of the \SN\ model}
\label{tbl:exs}
\begin{center}
\begin{tabular}{l l l}
Expression & Value & Type\\ 
\hline
$\mt{as\!:\!U}$ & $\mt{as\!:\!U}$ & binding \\
$\mt{qf\!:\!N \? N}$ & $\mt{qf}$ & unbind \\
$\mt{qf\!:\!N \? X}$ & $\mt{\$}$ & unbind (not found) \\
$\mt{(ao\!:\!N \+ ax\!:\!F \+ wh\!:\!A \? F)\!:\!K}$ & $\mt{ax\!:\!K}$ & 3-bind, unbind, rebind \\
$\mt{( ( ( ( sf\!:\!W \+ fr\!:\!V )\!:\!N )\!:\!R )\!:\!R )\!:\!Y \? Y \? R \? R \? N}$ & $\mt{sf\!:\!W \+ fr\!: \!V}$ & 4-nested, unbind \\
\hline
\end{tabular}
\end{center}
\end{table}

\section{The \SN\ model and experimental results}
\label{sec:model}
\SN\ is a standard bidirectional encoder-decoder network where the output of the bidirectional LSTM encoder is the \SR\ embedding of the input \SL\ expression. 
The \SR\ vector is then fed as input to an LSTM decoder. 
Some implementation details are given in Table~\ref{tbl:results}, 
which also gives the results of training \SN\ on randomly-selected input/output pairs.
\label{sec:results}

\begin{table}[t]
\vspace{-.4cm}
\caption{Performance and hyperparameters of the trained \SN\ model}
\label{tbl:results}
\begin{center}
\begin{small}
\begin{tabular}{llllllllll}
Accu- & Test per-& Train & Mini- & Batch & Hidden & Drop- & Learn- & Optim- & Attention; \\
racy & plexity & loss & batches & size & dimension & out  & ing rate & izer & Beam \\
\hline \\
96.16 & 1.02 & 0.187 & 54K & 128 & 128 & 0.2 & 0.001 & Adam & None \\
\hline
\vspace{-1cm}
\end{tabular}
\end{small}
\end{center}
\end{table}

\section{Analysis of \SR: The Superposition Principle}
\label{sec:superpos}

\subsectionPH{The Superposition Principle in theoretical structure-embedding schemes}
\label{sec:superTheo}
Theoretical solutions to performing the task defined in Sec.~\ref{sec:task} were proposed in the previous generation of neural network modeling.
Two general solutions, TPR and HRR, were introduced in Sec.~\ref{sec:overview}.
The TPR embedding of a symbol structure $\mc{S}$ with symbols $\{\mt{s_{k}}\}$ respectively bound to roles $\{\mt{r_{k}}\}$ is $\mr{TPR}(\mc{S}) \equiv \sum_{k} \mb{s_{k}} \otimes \mb{r_{k}}$, where $\otimes$ denotes the tensor (generalized outer) product and $\{\mb{s_{k}}\}$ and $\{\mb{r_{k}}\}$ are embeddings of the symbols and roles, with respective dimensions $\sigma$ and $\rho$; the dimension of the TPR itself is then $\sigma \rho$.

\ignore{
Recursive roles are defined via the tensor product as well, so in the binary tree example of Sec.~\ref{sec:expr}, $\mb{R_{01}} \equiv \mb{R_{0}} \otimes \mb{R_{1}}$.
Thus a symbol at depth $d$ in the tree is embedded as a rank-($d+1$) tensor, and the sum in the definition of a TPR is then a direct sum over vector spaces of dimension $\sigma \rho_{0}^{d}$, where $\rho_{0}$ is the dimension of the embeddings of $\mt{R_{0}}, \mt{R_{1}}$.
}
If the role-embedding vectors $\{\mb{r_{k}}\}$ are linearly independent, when collected together they form an invertible matrix $\mb{R}$; the rows of $\mb{R}^{-1}$ are the ``unbinding'' vectors $\{\mb{u_{k}}\}$: $\mb{r_{k}} \cdot \mb{u_{j}} = \delta_{kj}$ so these vectors can be used to unbind the roles in a TPR.
The symbol that fills role $\mt{r_{k}}$ in structure $\mc{S}$ is exactly the symbol $\mt{s_{k}}$ with embedding $\mb{s_{k}} = \mr{TPR}(\mc{S}) \cdot \mb{u_{k}}$.
\ignore{
This equation is exact.
However the requirement that the role-embedding vectors be linearly independent entails that the dimension of the role embedding, $\rho$, must be at least the number of distinct roles.
In the case of binary trees, two roles that are used recursively to generate all roles, so their embedding dimension $\rho_{0}$ must be at least 2. 
}

A crucial property of TPR is that the embedding of a structure is the \emph{sum} over embeddings of its symbols. This is TPR's Superposition Principle.
This is what enables extraction of symbols from any binding: $(\sum_{k} \mb{s_{k}} \otimes \mb{r_{k}}) \cdot \mb{u_{j}} = \sum_{k} \mb{s_{k}} \otimes (\mb{r_{k}} \cdot \mb{u_{j}}) = \mb{s_{j}}$ since $\mb{r_{k}} \cdot \mb{u_{j}} = \delta_{kj}$.

HRRs are essentially contracted TPRs \citep[260]{Smolensky06harmonic}. 
The equation defining TPR($\mc{S}$) also defines HRR($\mc{S}$), 
provided $\otimes$ is reinterpreted to denote circular convolution: 
$[\mb{a} \otimes \mb{b}]_{\mu} = \sum_{\nu} [\mb{a}]_{\nu} [\mb{b}]_{\mu - \nu}$.
Assuming the elements of the $\{\mb{r_{k}}\}$ are randomly (typically, normally) distributed, each role-embedding vector $\mb{r_{k}}$ can be used as its own unbinding vector.
However the HRR unbinding equation holds only approximately: $\mb{s_{k}} = \mr{HRR}(\mc{S}) \cdot \mb{u_{k}} + \mr{noise}$.
This noise must be eliminated by `clean-up' processes.
Note that, like TPR, HRR obeys the Superposition Principle.

\subsectionPH{Testing the Superposition Principle in the learned representation}
\label{sec:superModel}
As a test of whether the Superposition Principle holds of \SR, let $\mb{v}(\mt{expr})$ denote the \SR\ vector embedding of \SL\ expression $\mt{expr}$, and consider expressions containing two symbol/role bindings, such as $\mt{aa\!:\!A \+ bb\!:\!B}$.
Then if the Superposition Principle holds, we have\footnote{The simpler equation $\mb{v}(\mt{aa\!:\!A \+ bb\!:\!B}) - [\mb{v}(\mt{aa\!:\!A}) + \mb{v}(\mt{bb\!:\!B})] = \mb{0}$ does not hold in \SR; it appears that the manifolds of one- and two-binding embeddings are distinct. 
Eq.~\ref{eq:superSmall} is designed to stay within the latter.
Eq. \ref{eq:superSmall} is analogous to the famous equation 
$[\bv(\mr{king}) - \bv(\mr{man})] - [\bv(\mr{queen}) - \bv(\mr{woman})] \approx \bf{0}$
of \citet{Mikolov13}.
To make the analogy exact, let the roles `gender', `status' be denoted $\mt{G}, \mt{S}$ and let $\mt{male}, \mt{female}, \mt{royal}, \mt{commoner}$ be denoted $\mt{mm}, \mt{ff}, \mt{rr}, \mt{cc}$.
Then 
$\bf{0} \approx [\bv(\mr{king}) - \bv(\mr{man})] - [\bv(\mr{queen}) - \bv(\mr{woman})]$
$= [\bv(\mt{mm\!:\!G} \+ \mt{rr\!:\!S}) -\bv(\mt{mm\!:\!G} \+ \mt{cc\!:\!S})] - 
[\bv(\mt{ff\!:\!G} \+ \mt{rr\!:\!S}) - \bv(\mt{ff\!:\!G} \+ \mt{cc\!:\!S})$, 
i.e., Eq.~\ref{eq:superSmall}.
}:

\begin{enumerate}
\item 
$[\mb{v}(\mt{aa\!:\!A \+ bb\!:\!B}) - \mb{v}(\mt{aa\!:\!A \+ cc\!:\!C})] - [\mb{v}(\mt{dd\!:\!D \+ bb\!:\!B}) - \mb{v}(\mt{dd\!:\!D \+ cc\!:\!C})] = \mb{0}$. 
 \label{eq:superSmall}
\item 
$[\mb{v}(\mt{aa\!:\!A \+ bb\!:\!B}) - \mb{v}(\mt{aa\!:\!A \+ cc\!:\!C})] - [\mb{v}(\mt{xx\!:\!X \+ uu\!:\!U}) - \mb{v}(\mt{xx\!:\!X \+ vv\!:\!V})] \neq \mb{0}$. 
\label{eq:superBig}
\end{enumerate}

\begin{figure}[h]
\begin{center}
\vspace{-5cm}
\includegraphics[width=0.7\textwidth]{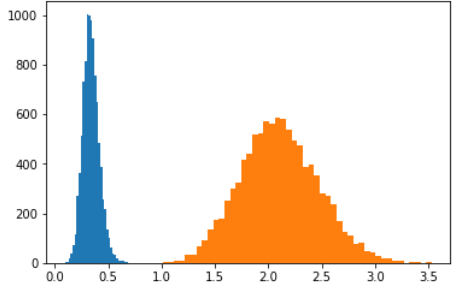}
\vspace{-5cm}
\end{center}
\caption{Distribution of lengths of vectors of the form of the LHS of Eqs.~\ref{eq:superSmall}  and \ref{eq:superBig}.}
\label{fig:super}
\end{figure}

We examine the Euclidean length of vectors of the form given on the LHS of Eq.~\ref{eq:superSmall}, which we expect to not be exactly 0, but small ---
small, for example, relative to the LHS of Eq.~\ref{eq:superBig}.
Fig.~\ref{fig:super} shows that this is true:
the AUC = 1.0 to within less than $10^{-16}$.
\\

\vspace{-.5cm}
\section{Conclusion}
A standard bidirectional encoder-decoder model can generate vector embeddings of expressions denoting complex symbol structures and can successfully query the content of such representations. 
Like theoretical techniques for accomplishing this, the learned representation obeys the Superposition Principle (approximately; at least within the manifold of embeddings of two-binding expressions).


\bibliography{iclr2018_workshop}

\begin{thebibliography}{18}
\providecommand{\natexlab}[1]{#1}
\providecommand{\url}[1]{\texttt{#1}}
\expandafter\ifx\csname urlstyle\endcsname\relax
  \providecommand{\doi}[1]{doi: #1}\else
  \providecommand{\doi}{doi: \begingroup \urlstyle{rm}\Url}\fi

\bibitem[Cho et~al.(2014)Cho, Merrienboer, Gulcehre, Bahdanau, Bougares,
  Schwenk, and Bengio]{Cho14}
Kyunghyun Cho, Bart~van Merrienboer, Caglar Gulcehre, Dzmitry Bahdanau, Fethi
  Bougares, Holger Schwenk, and Yoshua Bengio.
\newblock Learning phrase representations using rnn encoder-decoder for
  statistical machine translation.
\newblock \emph{Empirical Methods in Natural Language Processing-2014},
  abs/1406.1078, 2014.

\bibitem[Crawford et~al.(2016)Crawford, Gingerich, and Eliasmith]{Crawford16}
Eric Crawford, Matthew Gingerich, and Chris Eliasmith.
\newblock Biologically plausible, human-scale knowledge representation.
\newblock \emph{Cognitive Science}, 40\penalty0 (4):\penalty0 782--821, 2016.

\bibitem[Hinton(1988)]{Hinton88}
Geoffrey~E. Hinton.
\newblock Representing part-whole hierarchies in connectionist networks.
\newblock In \emph{Proceedings of the Tenth Annual Conference of the Cognitive
  Science Society}, pp.\  48--54. 1988.

\bibitem[Kanerva(2009)]{Kanerva09}
Pentti Kanerva.
\newblock Hyperdimensional computing: An introduction to computing in
  distributed representation with high-dimensional random vectors.
\newblock \emph{Cognitive Computing}, 1:\penalty0 139--159, 2009.

\bibitem[Lee et~al.(2016)Lee, He, Yih, Gao, Deng, and Smolensky]{Lee16}
Moontae Lee, Xiaodong He, Wen-tau Yih, Jianfeng Gao, Li~Deng, and Paul
  Smolensky.
\newblock Reasoning in vector space: An exploratory study of question
  answering.
\newblock In \emph{Proceedings of the International Conference on Learning
  Representations-2016}, 2016.

\bibitem[Mikolov et~al.(2013)Mikolov, Yih, and Zweig]{Mikolov13}
Tomas Mikolov, Scott Wen-tau Yih, and Geoffrey Zweig.
\newblock Linguistic regularities in continuous space word representations.
\newblock In \emph{Proceedings of the 2013 Conference of the North American
  Chapter of the Association for Computational Linguistics: Human Language
  Technologies (NAACL-HLT-2013)}. May 2013.

\bibitem[Newell(1980)]{Newell80physical}
Allen Newell.
\newblock Physical symbol systems.
\newblock \emph{Cognitive Science}, 4\penalty0 (1):\penalty0 135--183, 1980.

\bibitem[Palangi et~al.(2016)Palangi, Deng, Shen, Gao, He, Chen, Song, and
  Ward]{Palangi16deep}
Hamid Palangi, Li~Deng, Yelong Shen, Jianfeng Gao, Xiaodong He, Jianshu Chen,
  Xinying Song, and Rabab Ward.
\newblock Deep sentence embedding using long short-term memory networks:
  Analysis and application to information retrieval.
\newblock \emph{IEEE/ACM Transactions on Audio, Speech and Language Processing
  (TASLP)}, 24\penalty0 (4):\penalty0 694--707, 2016.

\bibitem[Plate(1993)]{Plate93holographic}
Tony Plate.
\newblock {Holographic Recurrent Networks}.
\newblock In Stephen~Jos{\'e} Hanson and C~Lee Giles (eds.), \emph{Advances in
  {Neural Information Processing Systems} 5}. Morgan Kaufmann, San Mateo, CA,
  1993.

\bibitem[Plate(2002)]{Plate02}
Tony Plate.
\newblock Distributed representations.
\newblock \emph{Encyclopedia of Cognitive Science}, 2002.

\bibitem[Plate(2003)]{Plate03holographic}
Tony Plate.
\newblock \emph{Holographic Reduced Representation: Distributed Representation
  for Cognitive Structures}.
\newblock CSLI Publications, Stanford, CA, 2003.

\bibitem[Pollack(1990)]{Pollack90recursive}
Jordan~B. Pollack.
\newblock Recursive distributed representations.
\newblock \emph{Artificial Intelligence}, 46\penalty0 (1):\penalty0 77--105,
  1990.

\bibitem[Smolensky(1990)]{Smolensky90tensor}
Paul Smolensky.
\newblock Tensor product variable binding and the representation of symbolic
  structures in connectionist networks.
\newblock \emph{Artificial Intelligence}, 46:\penalty0 159--216, 1990.

\bibitem[Smolensky \& Legendre(2006)Smolensky and
  Legendre]{Smolensky06harmonic}
Paul Smolensky and G{\'e}raldine Legendre.
\newblock \emph{The harmonic mind: From neural computation to
  {Optimality-Theoretic} grammar. 2 vols}.
\newblock MIT Press, Cambridge, MA, 2006.

\bibitem[Socher et~al.(2010)Socher, Manning, and Ng]{Socher10learning}
Richard Socher, Christopher~D Manning, and Andrew~Y Ng.
\newblock Learning continuous phrase representations and syntactic parsing with
  recursive neural networks.
\newblock In \emph{Proceedings of the NIPS-2010 Deep Learning and Unsupervised
  Feature Learning Workshop}, pp.\  1--9, 2010.

\bibitem[Touretzky(1990)]{Touretzky90}
David~S. Touretzky.
\newblock Boltzcons: Dynamic symbol structures in a connectionist network.
\newblock \emph{Artificial Intelligence}, 46:\penalty0 5--46, 1990.

\bibitem[Weston et~al.(2015)Weston, Bordes, Chopra, Rush, Merrienboer, Joulin,
  and Mikolov]{Weston15}
Jason Weston, Antoine Bordes, Sumit Chopra, Alexander~M. Rush, Bart~van
  Merrienboer, Armand Joulin, and Tomas Mikolov.
\newblock Towards {AI}-complete question answering: A set of prerequisite toy
  tasks.
\newblock \emph{arXiv:1502.05698}, 2015.

\bibitem[Zaremba \& Sutskever(2014)Zaremba and Sutskever]{Zaremba14}
Wojciech Zaremba and Ilya Sutskever.
\newblock Learning to execute.
\newblock \emph{arXiv:1410.4615}, 2014.

\end{thebibliography}
\bibliographystyle{iclr2018_workshop}

\end{document}